%% file: main.tex
\documentclass[10pt,twocolumn,letterpaper]{article}

\usepackage[accsupp]{axessibility}  % Improves PDF readability for those with disabilities.
\usepackage{iccv}
\usepackage{times}
\usepackage{epsfig}
\usepackage{graphicx}
\usepackage{amsmath}
\usepackage{amssymb}
\usepackage{booktabs}

\usepackage[pagebackref=true,breaklinks=true,letterpaper=true,colorlinks,bookmarks=false]{hyperref}

\usepackage{amsfonts}       % blackboard math symbols
\usepackage{nicefrac}       % compact symbols for 1/2, etc.
\usepackage{microtype}      % microtypography
\usepackage[table]{xcolor}        % colors

\usepackage{tikz}
\usepackage{import}
\usepackage{color}
\usepackage{multirow}
\usepackage{multicol}
\usepackage{adjustbox}
\usepackage{xspace}
\usepackage{wrapfig}
\usepackage{breqn}
\usepackage{enumitem}

\makeatletter
\DeclareRobustCommand\onedot{\futurelet\@let@token\@onedot}
\def\@onedot{\ifx\@let@token.\else.\null\fi\xspace}

\def\eg{\emph{e.g}\onedot} 
\def\ie{\emph{i.e}\onedot}

\def\etal{\emph{et al}\onedot}
\makeatother

\definecolor{darkcandyapplered}{rgb}{0.64, 0.0, 0.0}
\definecolor{burgundy}{rgb}{0.5, 0.0, 0.13}
\definecolor{carnelian}{rgb}{0.7, 0.11, 0.11}

\newcommand{\adrian}[1]{\textcolor{black}{#1}}\newcommand{\ric}[1]{\textcolor{black}{#1}}

\definecolor{tabhighlight}{HTML}{ffffff} %{ffcccb}
\definecolor{tabhighlightric}{HTML}{ffffff} %{cbccff}

\usepackage[capitalize]{cleveref}
\crefname{section}{Sec.}{Secs.}
\Crefname{section}{Section}{Sections}
\Crefname{table}{Table}{Tables}
\crefname{table}{Tab.}{Tabs.}

\iccvfinalcopy % *** Uncomment this line for the final submission

\ificcvfinal\fi

\def\mname{FS-DETR}

\newcommand*{\affaddr}[1]{#1} % No op here. Customize it for different styles.
\newcommand*{\affmark}[1][*]{\textsuperscript{#1}}

\begin{document}

\title{FS-DETR: Few-Shot DEtection TRansformer with prompting and without re-training}

\author{%
Adrian Bulat\affmark[1,2], Ricardo Guerrero\affmark[1], Brais Martinez\affmark[1], Georgios Tzimiropoulos\affmark[1,3]\\
\affaddr{\affmark[1]Samsung AI Cambridge}\;\;
\affaddr{\affmark[2]Technical University of Iasi}\;\;
\affaddr{\affmark[3]Queen Mary University of London}
}
  
\maketitle

\begin{abstract}
This paper is on Few-Shot Object Detection (FSOD), where given a few templates (examples) depicting a \textit{novel} class (not seen during training), the goal is to detect all of its occurrences within a set of images. From a practical perspective, an FSOD system must fulfil the following \textit{desiderata}: (a) it must be used as is, without requiring \textit{any} fine-tuning at test time, (b) it must be able to process an arbitrary number of novel objects concurrently while supporting an arbitrary number of examples from each class and (c) it must achieve accuracy comparable to a closed system. Towards satisfying (a)-(c), in this work, we make the following contributions: We introduce, for the first time, a simple, yet powerful, few-shot detection transformer (FS-DETR) based on visual prompting that can address both desiderata (a) and (b). Our system builds upon the DETR framework, extending it based on two key ideas: (1) feed the provided visual templates of the novel classes as visual prompts during test time, and (2) ``stamp'' these prompts with pseudo-class embeddings (akin to soft prompting), which are then predicted at the output of the decoder. Importantly, we show that our system is not only more flexible than existing methods, but also, it makes a step towards satisfying desideratum (c). Specifically, it is significantly more accurate than all methods that do not require fine-tuning and even matches and outperforms the current state-of-the-art fine-tuning based methods on the most well-established benchmarks (PASCAL VOC \& MSCOCO).
\end{abstract}

\vspace{-1em}

\section{Introduction}

\import{sections/}{introduction}

\section{Related work}
  \import{sections/}{related_work}

\section{Method}\label{sec:detr}

\import{sections/}{method}

\section{Experiments}\label{sec:experiments}
    \import{sections/}{experiments}

\section{Ablation studies}\label{sec:ablation}
    \import{sections/}{ablation}

\vspace{-1em}
\section{Conclusions}
    \import{sections/}{conclusions}

% \clearpage

%%%%%%%%% REFERENCES
{\small
\bibliographystyle{ieee_fullname}
\bibliography{egbib}
}

\appendix

\section{ Implementations details}\label{sec:supp-impl}

\mname~ extends Conditional DETR~\cite{meng2021conditional} (see Section~3), and was pre-trained and trained on 
a single node with
8 P40 GPUs. Following~\cite{dai2021up}, the ResNet50~\cite{he2016deep} backbone is initialized from SwAV~\cite{caron2020unsupervised} and kept frozen. Pre-training makes use of ImageNet-100 without labels, with object proposals detection as a pretext task. 

\textit{Pre-training hyper-parameters} were set to: Batch size of 32 per GPU, AdamW optimizer~\cite{loshchilov2017decoupled} with a learning rate of $10^{-4}$, frozen backbone CNN, path dropout of 0.1, training for 60 epochs with the learning rate decreased by factor of $10$ after $40$ epochs. When using larger images for pre-training (\ie containing complex scenes) the batch size is decreased to 2.

\textit{Training hyper-parameters} were set to: Batch size of 2 per GPU, SGD with momentum (0.9)~\cite{qian1999momentum} with the learning rate initially set to $5e^{-1}$, path dropout of $0.1$, training for 30 epochs with the learning rate decreased by a factor of 10 after 20 epochs (and respectively 15 for COCO). Augmentation followed DETR: input images were resized such that the short axis is 480 at least and 800 pixels at most, and the long side is, at most, 1333 pixels, and randomly cropped with $0.5$ probability.

\textit{Patch augmentation hyper-parameters.} The templates are cropped tightly based on the bounding box and then rescaled to a $128\times 128$px image. During training we apply the following augmentations: color jittering, with 0.8 probability and 0.4 intensity, random gray scale (0.2 probability) and Gaussian blur with a probability of 0.5. 

\section{Pre-training process} Transformer based architectures are known to generally be more data-hungry than their homologous CNNs~\cite{dosovitskiy2020image,carion2020end}. To alleviate this, we introduce a label-free pre-training step that closely mimics the training stage.

\begin{figure*}[t]
    \centering
        \includegraphics[width=12cm, trim={0.2cm 0.2cm 0.2cm 0.2cm},clip]{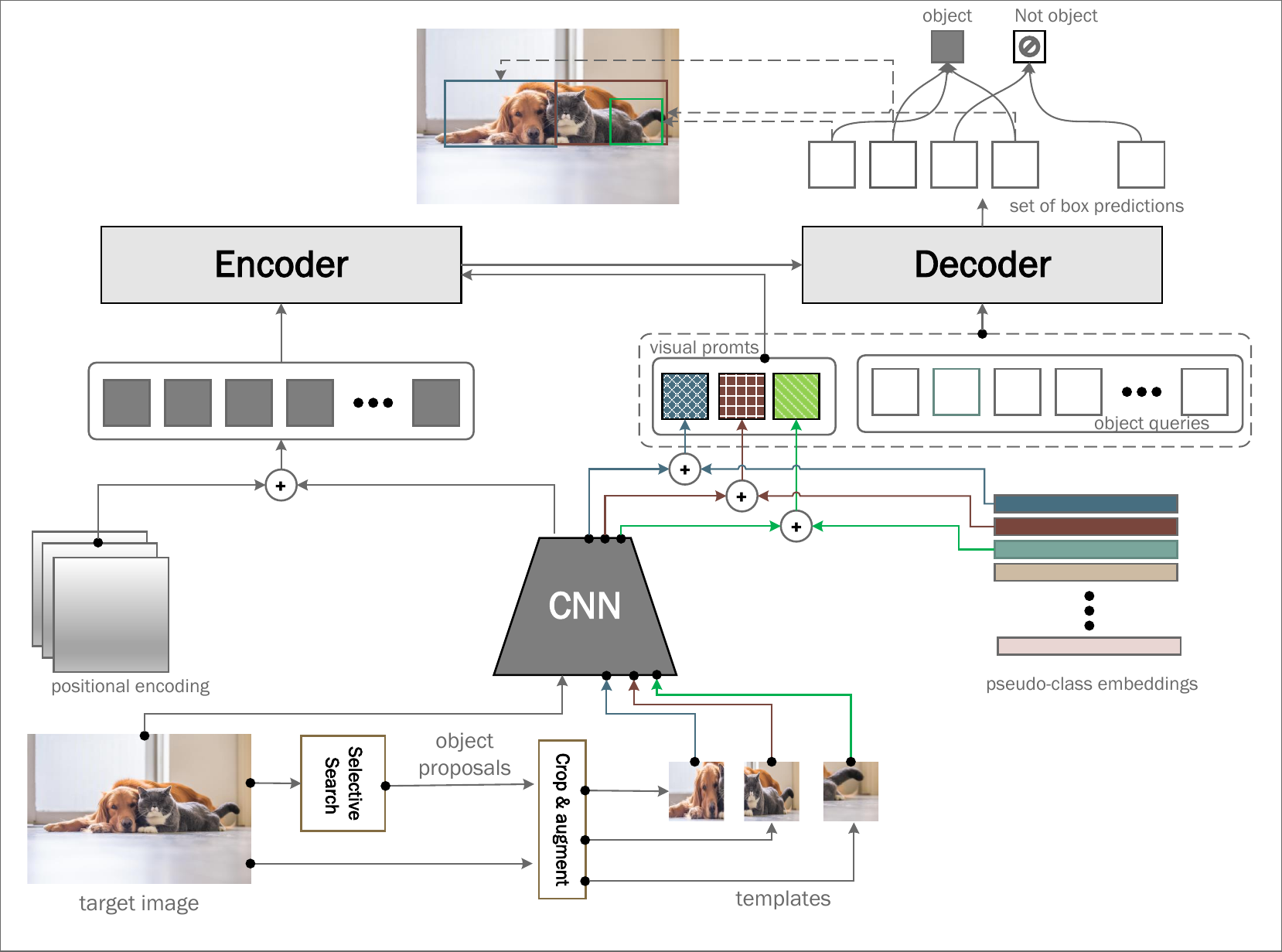}
    \caption{\mname~ pre-training stage. The pre-training process largely mimics the training stage, with a few notable differences: (1) no annotations are used, (2) the target bounding boxes are proposed by selective search or sampled randomly, (3) the templates are sampled from the target image itself and (4) only two classes are defined - object and no object.}
    \label{fig:pretraining}
\end{figure*}

More specifically, at train time, for any given input image, we crop a set of patches according to the object proposals produced by Selective Search~\cite{uijlings2013selective}~\footnote{ 
Selective Search is a  \textbf{training-free} generic region proposal algorithm that computes a hierarchical grouping of image regions based on color, texture, size and shape, and hence, has no notion of object classes.
}. Each of these patches represents an object (belonging to some class) and can be mapped to a pseudo-class, by associating it to a different pseudo-class embedding. Note, that random patches can be used too, but the former leads to faster convergence.
The goal of the network is to predict the location of these patches (\ie object templates). To make the task harder, the patches (templates) are augmented using a set of random transformations before being passed to the backbone. Finally, the network is trained using a regression (for the bounding boxes) and a classification loss. As opposed to the supervised training stage, the classification loss is reduced to a binary classification problem \adrian{initially}: object/no object \adrian{and then to the proposed loss, after this warm-up}. The model is then trained using the hyper-parameters described in Section~\ref{sec:supp-impl} while the ResNet based backone is initialised from a model pre-trained on Imagenet without supervision (SwAV~\cite{caron2020unsupervised}). \adrian{Note that unlike~\cite{dai2021up,bar2022detreg} that also make use of unsupervised detection-centric training, our work concatentes a set of templates as prompts, instead of grouped-based summation, uses a differnt trainign objective and makes use of negative templates.} The process is illustrated in Fig.~\ref{fig:pretraining}.

\noindent\textbf{Pre-training dataset} 
For our DETR pre-training, we used the images belonging to the base classes from COCO (60 classes in total) and ImageNet-100 (a  subset of ImageNet introduced in~\cite{tian2019contrastive}). We note the following: firstly, there is no overlap between COCO base classes and VOC and COCO novel classes. Secondly, ImageNet-100 contains classes that can be matched to 7 out of 20 VOC classes (bird, cat, dog, boat, car, motorcycle and chair). Specifically, split-1 of VOC novel classes contains 2/5 classes (bird and motorbike) that overlap with ImageNet-100, split-2 0/5 and split-3 3/5 (boat, cat and  motorbike). Please note that NONE of the labels in ImageNet-100 (or COCO) are used at any stage of the pre-training. While we agree that the underlying data distribution, even for unsupervised learning is important, judging from the results from Tables 1 and 2 the gains in absolute terms offered by our approach are consistent across all 3 sets (note that split-2 has no overlap at all).

We note that, recent state-of-the-art methods (\eg Fan et al~\cite{fan2020few}, QA-FewShot~\cite{han2021query}, DeFRCN~\cite{qiao2021defrcn}) make use of a backbone pre-trained with full supervision on the entire ImageNet, same which includes all VOC/COCO novel classes. In this regard, we trained \mname~ initialized from a backbone pre-trained on the entirety of Imagenet for classification using full supervision (\eg. same as~\cite{fan2020few,han2021query,qiao2021defrcn}). Preliminary results shown in Tab.~\ref{tab:intializations} (which could likely be improved from hyper-parameter optimization) indicate an overall improvement of approx. 1.5\%. This highlights that the pre-training data used in the proposed work doesn't offer any advantage over prior art approaches that use fully supervised pre-trained backbones. Further to this, DeFRCN~\cite{qiao2021defrcn} experimented with using a backbone pre-trained on ImageNet without labels (SwAV weights - same as ours) which resulted in substantially degraded performance of approx. 5.0\%.

\begin{table}[ht]
\centering
\footnotesize
\setlength{\tabcolsep}{0.4em}
% \vspace{-0.1cm}
\caption{Impact of different initialisation of backbone on the PASCAL VOC dataset (Novel Set 1). }
%\adjustbox{width=0.6\linewidth}{
\begin{tabular}{c|ccccc}
\toprule
\multirow{2}{*}{Approach}  &  \multicolumn{5}{c}{Novel Set 1} \\ 
&   1     & 2     & 3    & 5    & 10     \\  \midrule
\mname~ (Swav)	& 45.0	& 48.5	& 51.5	& 52.7	& 56.1    \\
\mname~ (ImageNet) & 47.1	& 49.9	& 52.5	& 53.8	& 57.0   \\
\bottomrule
\end{tabular}%}
\label{tab:intializations}
%\vspace{-0.25cm}
\end{table}

\section{Qualitative evaluation}

\begin{figure}[t]
    \centering
    \includegraphics[width=\linewidth]{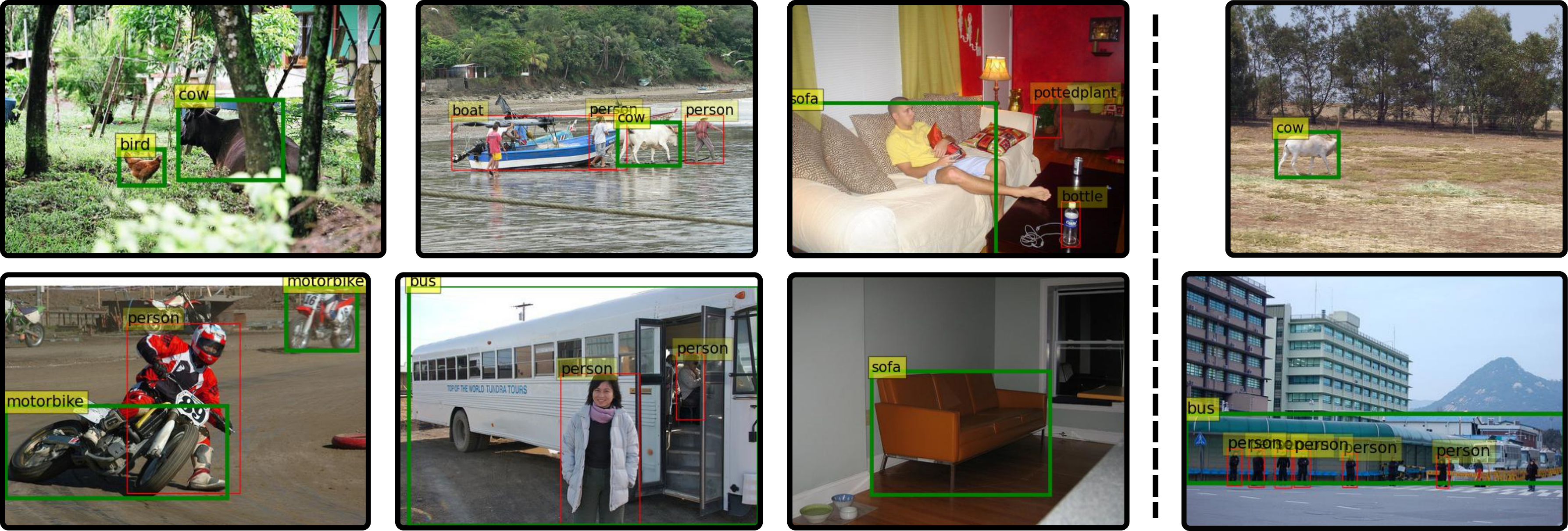}
    \caption{
    Novel class 1-shot detection examples with \mname. First three columns depict success cases, while the right-most column failures. Green and red boxes indicate novel and base classes, respectively. Note that in the top-left image two novel classes are detected simultaneously.
    }
    \label{fig:success_fail}
    \vspace{-0.5cm}
\end{figure}

Fig.~\ref{fig:success_fail} shows 1-shot detection examples of \mname~, with success cases shown on the first three columns, and fail cases on the right-most column. The image on top-left of the figure, illustrates an important and unique property of \mname~: Two novel classes coexist in a single image, and \mname~ is able to successfully detect both of them at the same time.

\begin{figure*}[!ht]
    \centering
    \includegraphics[width=\linewidth]{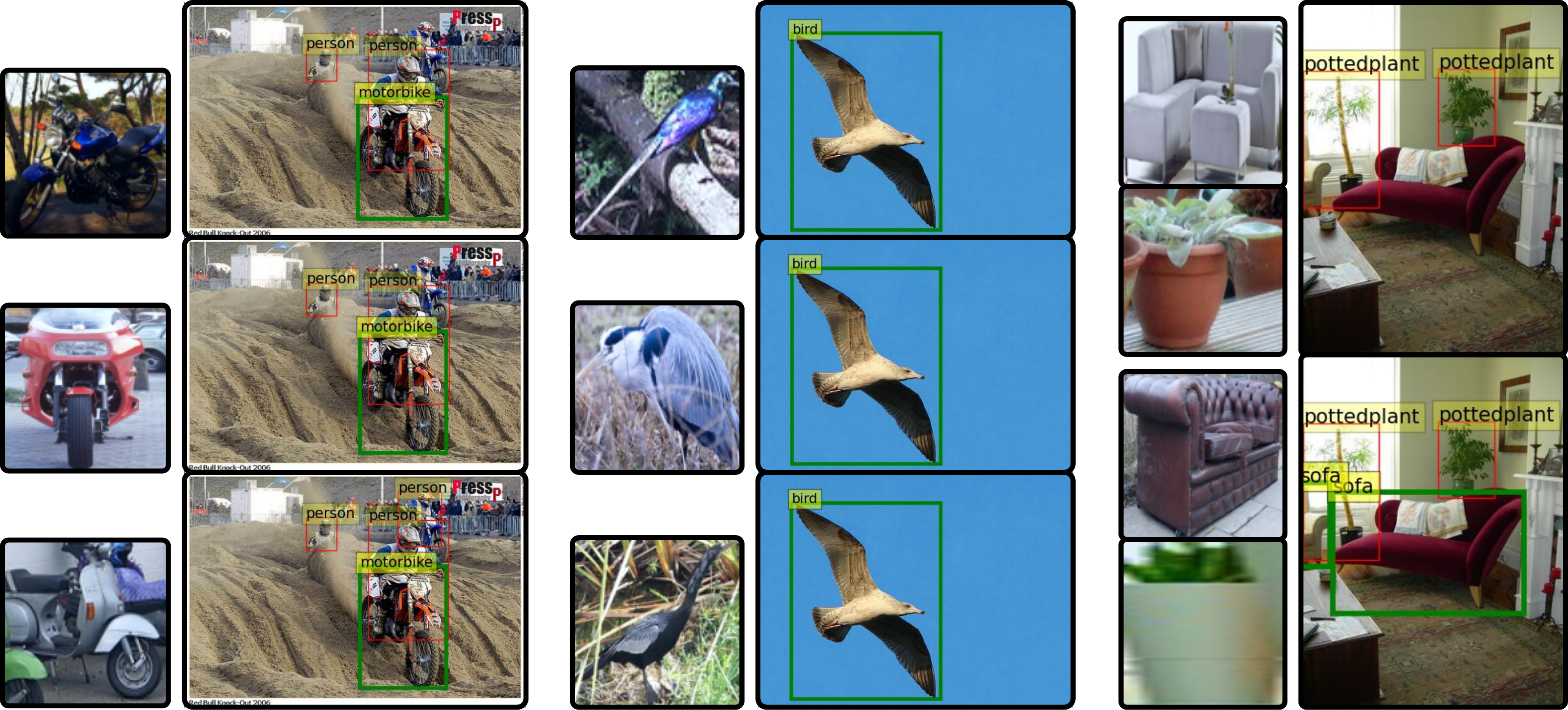}
    \caption{Effect of different 1-shot template on detection with \mname. Small images indicate the
    % 1-shot 
    template used to detect the objects on the larger images. The left-most two pairs of columns illustrate the robustness to template change, while the right-most column pair illustrates a failure case.
    }
    \label{fig:different_template}
    \vspace{-0.5cm}
\end{figure*}

Fig.~\ref{fig:different_template} shows the effect of varying the 1-shot template used during novel class detection. There, smaller images refer to the templates used for 1-shot detection on the paired larger image. From the left-most two pairs of columns, it can be appreciated that even under large template visual variability, \mname~ proves to be extremely robust, with detections hardly affected by the template change. The right-most illustrates a failure case, where the sofa fails to be detected.

\begin{figure*}[t]
    \centering
    \vspace{0.25cm}
        \includegraphics[width=13cm, trim={0.2cm 0.2cm 0.2cm 0.2cm},clip]{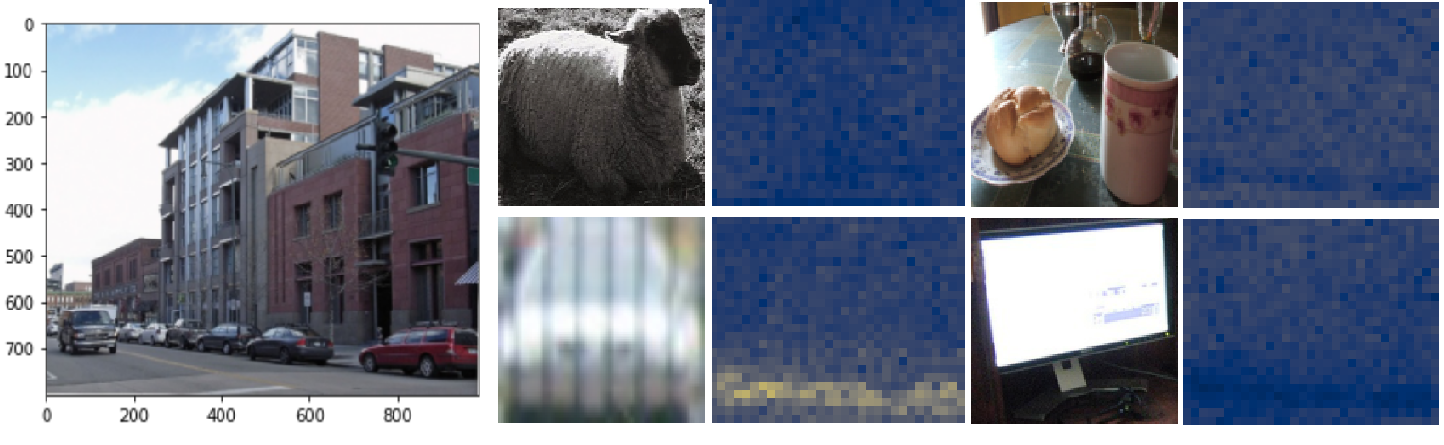} \\
        \vspace{0.25cm}
        \includegraphics[width=13cm, trim={0.2cm 0.2cm 0.2cm 0.2cm},clip]{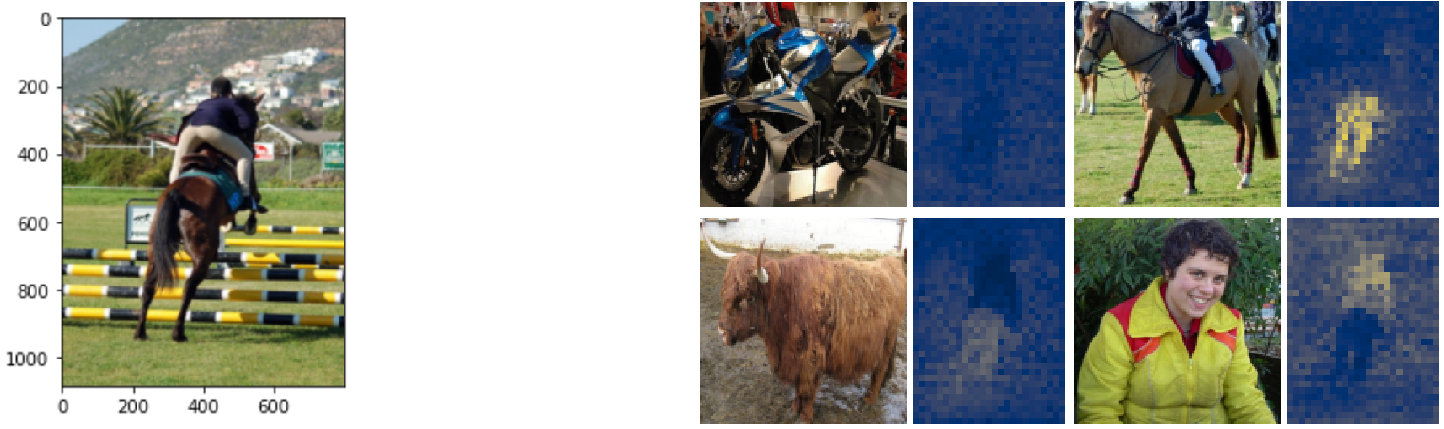} \\
        \vspace{0.25cm}
        \includegraphics[width=13cm, trim={0.2cm 0.2cm 0.2cm 0.2cm},clip]{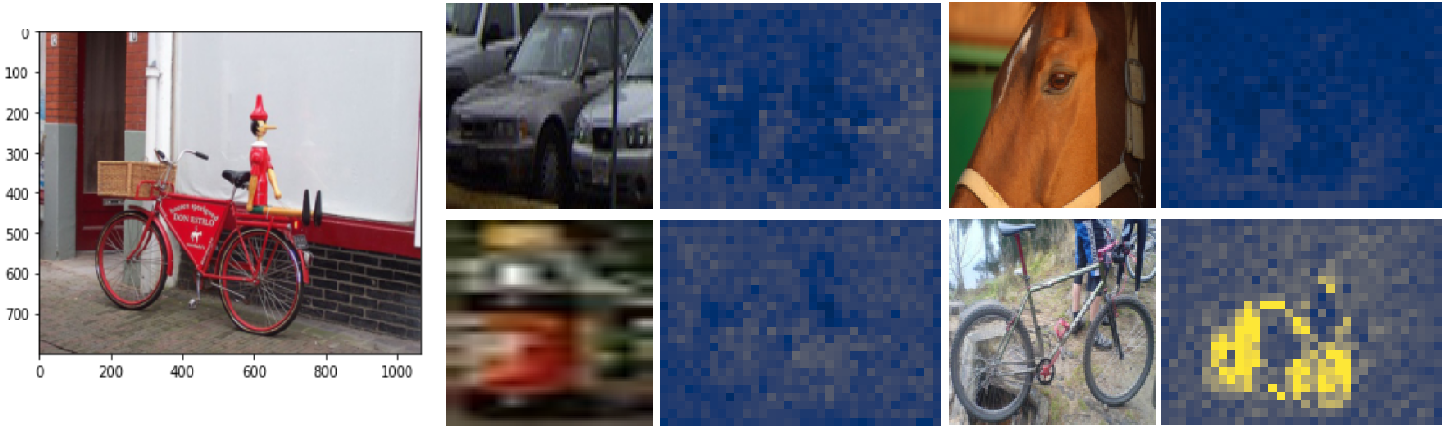}
    \caption{Attention weights between the visual prompts (templates) and the encoded image features for three randomly sampled target images (left column) from VOC Pascal dataset. Notice that the network learns to attend to the parts of the image that are semantically close to the presented templates. 
    % The images on the left are the target images. 
    For each target image (left column), we show the attention weights generated by four templates. We observe that for the target image of the first row, only the car template generates attention of high magnitude at several locations corresponding to the location of the cars in the target image. Similarly, for the target image of the second row only the horse and the person templates fire at the corresponding locations in the target image as expected. Similar conclusions can be drawn for the target image of the last row.}
    \label{fig:attn_maps}
    \vspace{-0.25cm}
\end{figure*}

Additionally, in Fig.~\ref{fig:attn_maps} we visualise the attention weights between the visual prompts and the encoded image features. Notice that our network learns to attend to parts of the target image that are semantically similar to the provided templates that are present in the target image.

\section{Discussion, challenges and limitations}

Herein we offer a pertinent discussion on some things we tried but didn't work, defining some of the limitations and challenges that arise within the proposed framework and more so in general for FSOD using images within DETR framework.

\subsection{Few-shot object detection objective ambiguity}

A general limitation of few shot object recognition systems, trained and/or tested using one or more visual examples is the ill-definess of what represents a class. For example, presenting a template depicting a dog could require identifying the class ``dog'', ``bulldog''(\ie find dogs of a given bred), ``a white dog'' etc. While as the number of examples increases the ambiguity decreases, the problem is not fully solvable within the visual domain. A natural solution to this problem could be provided by constructing the templates using natural language. While an interesting solution, this goes beyond the scope of this work.

That being said, to some extent, our approach alleviates parts of this problem: As our model has to distinguish locally within the set of provided positive (present in the image) and negative (not present) templates, it can use them to semantically ground the notion of a class, effectively defining the semantic hierarchy. For example, if all templates are representing different apple varieties, the model is expected to differentiate between these varieties instead of detecting \textit{any} apple.

\subsection{Challenges within the DETR framework} Despite its remarkable success and appealing formulation that removes the need of an explicit object proposal component or post-processing step (\ie NMS), in the context of few-shot detection some of this advantages pose additional challenges, some of which we detail bellow. We believe this aspects could represent potentially interesting future exploration directions.

\noindent\textbf{Semantic misalignment} Traditional object detection systems, such as~\cite{ren2015faster,redmon2016you,he2017mask} preserve an exact feature alignment between the regressed bounding box and the semantic information (\ie the ROI pooling extracts features at the location given by the proposal). DETR derived approaches however construct their representation gradually by adapting a set of object queries via self-attention and cross-attention with the encoded features. As each object query operates (attends) to the entire image, as opposed to the local ROI, the query can encode information outside of the predicted bounding box. Thus, we can get to cases where the class may be correct although the bounding box contains mostly objects of an incorrect category.

Therefore, when we tried to use an external supervised classifier, applied to the image region cropped based on the predicted bounding box, surprisingly we noticed a deterioration of the performance. Upon visual inspection we observed a manifestation of the above mentioned phenomena, where the model was able to predict the correct class despite the fact that the predicted bounding box contained predominantly content of a different class, while the external supervised classifier was unable to.

\noindent\textbf{Reduced proposal diversity} A key characteristic of DETR systems is the removal of an a) external object proposal generator and b) implicit Non Maximum Suppression (NMS). Upon close inspection of our system we noticed that as we advance within the transformer based decoder, the bounding boxes are pruned via self-attention. By the end, despite having 100-300 object queries, most will point to a very small set of distinct regions of the image, lacking the diversity present in more traditional systems, such as in Fast RCNN.  The consequence of this is a higher likelihood of missing unseen classes in limited data scenarios, making the pre-training even more so important to train the built-in object proposals system.

\end{document}

%% file: sections/introduction.tex
Thanks to the advent of deep learning, object detection has witnessed tremendous progress over the last years. However, the standard setting of training and testing on a closed set of classes has specific important limitations. Firstly, it's unfeasible to annotate all objects of relevance present in-the-wild, thus, current systems are trained only on a small subset. It does not seem straightforward to significantly scale up this figure. Secondly, human perception operates mostly under the open set recognition/detection setting. Humans can detect/track new unseen objects on the fly, typically using a single template, without requiring any ``re-training'' or ``fine-tuning'' of their ``detection'' skills, arguably a consequence of the prior representation learned, an aspect we sought to exploit here too. Finally, important applications in robotics, where agents may interact with previously unseen objects, might require their subsequent detection on the fly without any re-training. Few-Shot Object Detection (FSOD) refers to the problem of detecting a novel class not seen during training and, hence, can potentially address many of the aforementioned challenges.

There are still important \textit{desiderata} that current FSOD system must address in order to be practical and flexible to use: (a) They must be used as is, not requiring \textit{any} re-training (\eg fine-tuning) at test time \adrian{- a crucial component for autonomous exploration~\cite{li2022airdet}.} However, many existing state-of-the-art FSOD systems (\eg ~\cite{sun2021fsce,wu2021universal,qiao2021defrcn}) rely on re-training with the few available examples of the unseen classes. While such systems are still useful, the requirement for re-training makes them significantly more difficult to deploy on the fly and in real-time or on devices with limited capabilities for training. (b) They must be able to handle an arbitrary number of novel objects (and moreover an arbitrary number of examples per novel class) simultaneously during test time, in a single forward pass without requiring batching. This is akin to how closed systems work, which are able to detect multiple objects concurrently. However, to our knowledge there is no FSOD system  
possessing this property without requiring re-training. 
(c) They must attain classification accuracy that is comparable to that of closed systems. However, existing FSOD systems are far from achieving such high accuracy, especially for difficult datasets like MSCOCO.     

This work aims to significantly advance the state-of-the-art in all three above-mentioned challenges. To this end, and building upon the DETR~\cite{carion2020end} framework, we propose a system, called Few-Shot Detection Prompting (\mname), capable of detecting multiple novel classes at once, supporting a variable number of examples per class, and importantly, without any extra re-training. In our system, 
the visual template(s) (\ie prompts) from the new class(es) are used, during test time, in two ways: 
(1) in \mname's encoder to filter the backbone's image features via cross-attention, and more importantly, (2) as visual prompts in \mname's decoder, ``stamped'' with special pseudo-class encodings and prepended to the learnable object queries. The pseudo-class encodings are used as pseudo-classes which a classification head attached to the object queries is trained to predict via a Cross-Entropy loss. Finally, the output of the decoder are the predicted pseudo-classes and regressed bounding boxes. The two components, when combined allow the creation of a FSOD model that can localise, within one forward pass multiple objects at once, each with an arbitrary number of examples, without retraining.

\adrian{Contrary to prior work (\eg ~TSF~\cite{lai2022tsf} and AirDet~\cite{li2022airdet}), \mname, akin to soft-prompting~\cite{jia2022visual}, ``instructs'' the model in the input space regarding the visual appearance of the searched object(s). The network  is then capable of predicting for each prompt (\ie visual template) all the locations at which it is present in the image, if any. This is achieved without any additional modules or carefully engineered structures and feature filtering mechanisms (\eg  TSF~\cite{lai2022tsf} AirDet~\cite{li2022airdet}). Instead, we directly append the prompts to the object queries of the decoder. }

In summary, \textbf{our main contributions} are:

\begin{enumerate}
    \item
    We propose a fine-tuning-free Few-Shot Detection Prompting (\mname) method  which is capable of detecting multiple novel objects at once, and can support an arbitrary number of samples per class in an efficient manner via soft visual prompting.
    \item
    We show that all these features can be enabled by extending DETR based on two key ideas: (1) feed the provided visual templates of novel classes as visual prompts during test time, and (2) ``stamp'' these prompts with (class agnostic) pseudo-class embeddings, which are then predicted at the output of the decoder along with bounding boxes (akin to soft-prompting).
    \item 
    We also propose a simple and efficient yet powerful pipeline consisting of unsupervised pre-training followed by prompt-like base class training.
    \item 
    In addition to being more flexible, our system matches and outperforms state-of-the-art results on the standard FSOD setting on PASCAL VOC and MSCOCO. Specifically, \mname~ outperforms the not re-trained methods 
    of~\cite{han2021query,li2022airdet}
    and most re-training based methods on extreme few-shot settings ($k=1,2$), 
    while being competitive for more shots. 
\end{enumerate}

%% file: sections/related_work.tex
\noindent\textbf{DEtection TRansformer (DETR) approaches}: After revolutionizing NLP~\cite{vaswani2017attention,raffel2019exploring}, Transformer-based architectures have started making significant impact in computer vision problems~\cite{dosovitskiy2020image,liu2021swin}. In object detection, methods  are typically  grouped  into two-stage (proposal-based)~\cite{ren2015faster,he2017mask,cai2018cascade} and single-stage (proposal-free)\cite{lin2017focal,liu2016ssd,tian2019fcos,zhou2019objects,law2018cornernet} methods. In this field, a recent breakthrough is the DEtection TRansformer (DETR)~\cite{carion2020end}, which is a single-stage approach that treats the task as a direct set prediction without requiring hand-crafted components, like non-maximum suppression or anchor generation. Specifically, DETR is trained in an end-to-end manner using a set loss function which performs bipartite matching between the predicted and the ground-truth bounding boxes. Because DETR has slow training convergence, several methods have been proposed to improve it~\cite{meng2021conditional,zhu2020deformable,dai2021up}. Conditional DETR~\cite{meng2021conditional} learns a conditional  spatial  query  from  the decoder embeddings that are used in the decoder for cross-attention with the image features. Deformable DETR~\cite{zhu2020deformable} proposes deformable attention in which attention is performed only over a small set of key sampling points around a reference point. Unsupervised pre-training of DETR~\cite{dai2021up} (UP-DETR), improves its convergence, where randomly cropped patches are summed to the object queries and the model is then trained to detect them in the original image. \adrian{A follow-up work, DETReg~\cite{bar2022detreg}, replaces the random crops with proposals generated by Selective Search.}
While our approach is agnostic to the exact variant of DETR, 
due to its fast training convergence, we opted to use Conditional DETR as the model that we build our \mname~ approach upon. Beyond this, the above mentioned works are on closed set recognition and while UP-DETR's unsupervised pre-training could be potentially used for few-shot detection, the experimental setting presented in their work doesn't match the standard settings for few-shot detection and no code is provided for its training. We re-implemented UP-DETR~\cite{dai2021up} for few-shot detection and found that our method outperforms it. This is expected as their goal is unsupervised pre-training and not FSOD.

\noindent\textbf{Few Shot Object Detection (FSOD)} methods can be categorised into \textit{re-training based} and \textit{without re-training} methods. Re-training based methods assume that during test time, but before deployment, the provided samples of the novel categories can be used to fine-tune the model. This setting is restrictive as it requires training before deployment. Instead, without re-training methods can be directly deployed on the fly for the detection of novel examples. 

\noindent \textbf{\textit{Re-training based}} approaches can be divided into meta-learning and fine-tuning approaches. Meta-learning based approaches attempt to transfer knowledge from the base classes to the novel classes through meta-learning~\cite{finn2017model, gidaris2019generating, yan2019meta,wang2019meta, li2021transformation, xiao2020few}.
Fine-tuning based methods follow the standard pre-train and fine-tune pipeline. They have been shown to significantly outperform meta-learning approaches.
TFA~\cite{wang2020frustratingly} proposes fine-tuning the final classification layer of a Faster R-CNN model (first trained on base classes), with a balanced subset containing also the examples of the novel classes. SRR-FSD~\cite{zhu2021semantic} proposes to construct a semantic space using word embeddings, and then train a FSOD by projecting and aligning object visual features with their corresponding text embeddings. CME~\cite{li2021beyond} proposes to learn a feature embedding space where the margins between novel classes are maximised. Retentive R-CNN~\cite{fan2021generalized} addresses the problem of learning a FSOD without catastrophic forgetting (\ie without compromising base class accuracy). FSCE~\cite{sun2021fsce} aims to decrease instance similarity between 
objects belonging to different categories by adding a secondary branch to the primary RoI head, which is trained via supervised contrastive learning. 
The method of~\cite{zhang2021hallucination}
proposes a hallucinator network to generate 
examples which can help the classifier learn a better decision boundary for the novel classes. FSOD-UP~\cite{wu2021universal} proposes to construct universal prototypes capturing invariant object characteristics which, via fine-tuning, are adapted to the novel categories. DeFRCN~\cite{qiao2021defrcn}  proposes to perform stop-gradient between the RPN and the backbone, and scale-gradient between RCNN and the backbone. 

More recently, \ric{FSODMC~\cite{QiFan2022} proposes to address base class bias via novel class fine-tuning while calibrating the RPN, detector and backbone components to preserve well-learned prior knowledge.} \ric{KFSOD~\cite{zhang2022kfsod} improves upon \cite{fan2020few} by replacing the class-specific average-pooling of features
with kernel-pooled representations that are meant to capture non-linear patterns.}
\ric{TENET~\cite{zhang2022tenet} extends KFSOD with a multi-head attention transformer block on 
2nd-, 3rd- and 4th-order pooling.}
\ric{FCT~\cite{han2022few} extends~\cite{han2021query} by incorporating a cross-transformer into both the feature backbone and detection head 
to encourage 
query-support
multi-level interactions. Their approach is based on two-stage Faster-RCNN trained with a binary cross-entropy loss, \ie it is entirely different from our architecture and training objective based on pseudo-class prediction.} Meta-DETR~\cite{zhang2022meta} proposes a correlation aggregation module, which is then placed before a standard DETR encoder-decoder, that filters the query image tokens  using the support images and tasks. In contrast, we model the interactions directly via a novel visual template prompting formulation, without any additional modules and can process an arbitrary number of examples per-object and object within the same forward pass. Moreover, their method requires finetuning for FSOD deployment, while our doesn't require any retraining. 
\adrian{TSF~\cite{lai2022tsf} proposes a transformer plugin module for modelling interactions the input features $f$ and a set of learnable parameters $\theta$ representing base class information (\ie prototypes). In contrast to~\cite{lai2022tsf}, our approach does not learn any type of base class prototypes and is fully dynamic (interactions between data and data as opposed to data and prototypes)}.

\noindent\textbf{\textit{Without re-training}} approaches are primarily based on metric learning~\cite{vinyals2016matching,snell2017prototypical}. A standard approach is~\cite{hsieh2019one}, which uses cross-attention between the backbone's and the query's features to refine the proposal generation, then re-uses the query to re-weight the RoI features channel-wise (in a squeeze-and-excitation manner) for novel class classification. A similar approach for proposal generation is described in~\cite{fan2020few}, where the squeeze-and-excitation module is replaced with a multi-relation network. QA-FewDet~\cite{han2021query}  extends~\cite{hsieh2019one,fan2020few} 
by modelling class-class, class-proposal and proposal-proposal relationships using various GCNs.
\adrian{Finally, the concurrent work of AirDet~\cite{li2022airdet} attempts to learn a set of prototypes and a cross-scale support guided proposal network, with the association and regression performed at the 
end of the model via a detection head.}  To our knowledge, \adrian{AirDet} represents the state-of-the-art FSOD 
without re-training. We show that the proposed \mname~ outperforms it by a large margin.

\noindent\textbf{Relation to our work:} Our method is the first to perform re-training free visual prompting for few shot object detection.
Different to 
many other works (\eg TSF~\cite{lai2022tsf}, AirDet~\cite{li2022airdet}), \mname~ does not learn perform visual prompting nor learn class-related prototypes (\ie soft prompts-like). We emphasize that the pseudo-class embeddings in \mname~ are class-agnostic Finally, there are methods which are trained using metric learning~\cite{fan2020few, han2021query, han2022few} using a binary cross entropy loss. In contrast, \mname~ is trained to predict pseudo-classes using cross entropy (in a class-agnostic way) which is a more powerful training objective. 

%We do not propose 

\begin{figure*}[t]
    \centering
        \includegraphics[width=14cm]{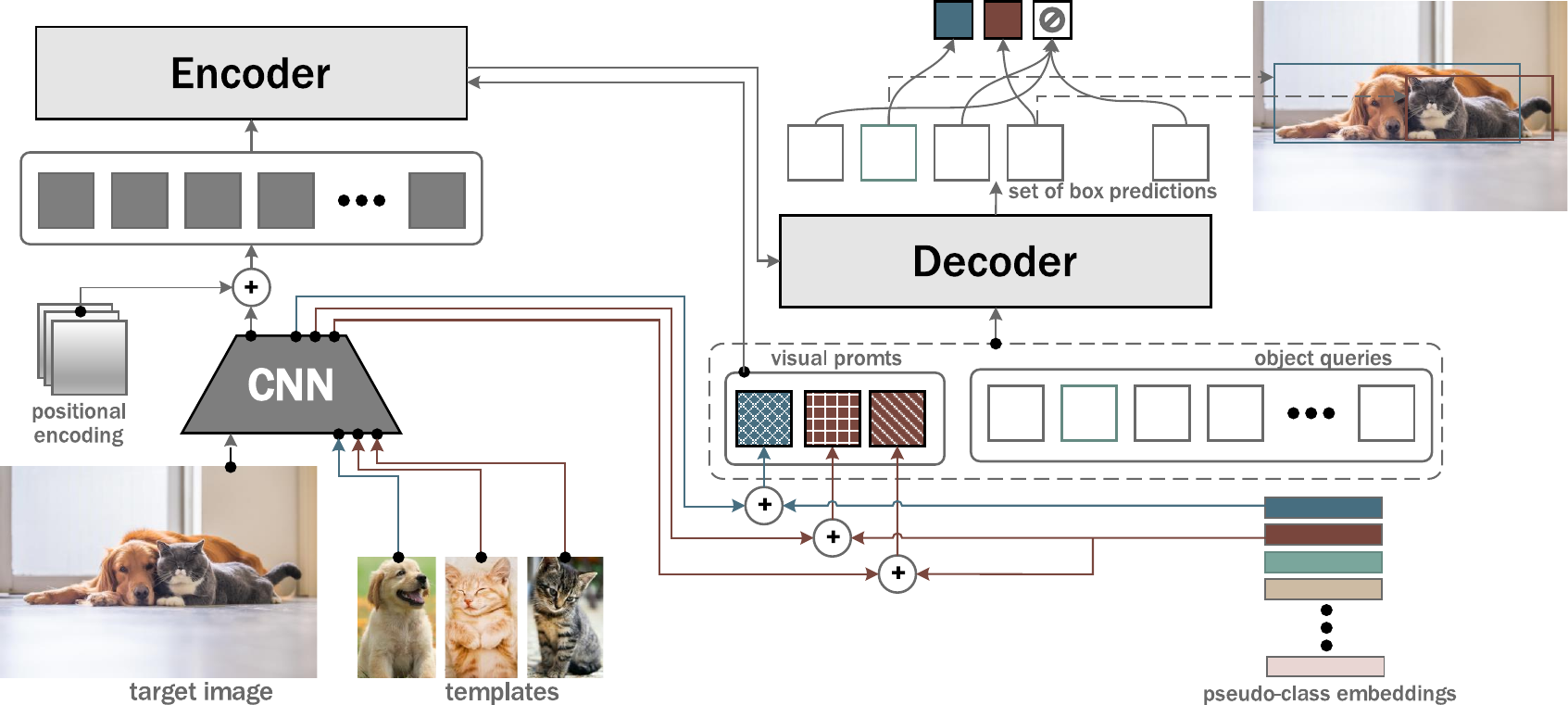}
    \caption{In the proposed \mname, the available templates are provided as additional \textit{visual prompts} to the system in order to condition 
    % upon 
    and control the output.
    % of the detector. 
    To train and test the system, these prompts are ``stamped'' with \textit{pseudo-class embeddings} (see Sec.~\ref{ssec:\mname-arch}) which are predicted at the output of the decoder along with bounding boxes (note, that there is \textit{no correlation} between actual classes and pseudo-classes, \eg the cat could be of either class: ``blue'' or ``red`` as there is no preferred order). \mname~ naturally supports $k-$shot detection, as the model can process multiple examples per class at once. Templates belonging to the same class will share the same pseudo-class embedding. Red and blue colors denote the different pseudo-classes associated to the input templates.}
    \label{fig:overall_idea}
    \vspace{-0.5cm}
\end{figure*}

%% file: sections/method.tex
Given a dataset where each image is annotated with a set of bounding boxes representing the instantiations of $C$ known base classes, our goal is to train a model capable of localizing objects belonging to novel classes, \ie unseen during training, using up to $k$ examples per novel class. In practice, we partition the available datasets into two disjoint sets, one containing $C_{novel}$ classes for testing, and another with $C_{base}$ classes for training (\ie $C = C_{novel} \cup C_{base}$ and $C_{novel} \cap C_{base}=\emptyset$). 

\subsection{Overview of \mname} 

We build the proposed Few-Shot DEtection TRansformer (\mname) upon DETR's architecture~\footnote{We note that, in practice, due to its superior convergence properties, we used the Conditional DETR as the basis of our implementation but for simplicity of exposition we will use the original DETR architecture.}. \mname's architecture consists of: (1) the CNN backbone used to extract visual features from the target image and the templates, (2) a transformer encoder that performs self-attention on the image tokens and cross-attention between the templates and the image tokens, and (3) a transformer decoder that processes object queries and templates to make predictions for pseudo-classes (see also below) and bounding boxes. Contrary to the related works of~\cite{fan2020few,han2021query,han2022meta}, our system processes an arbitrary number of templates (\ie new classes) jointly, in a single forward pass, \ie without requiring batching, significantly improving the efficiency of the process.

\noindent \textbf{Key contributions:} DETR re-formulates object detection as a set prediction problem via prompting, making object predictions by ``tuning'' a set of $N$ learnable queries $\mathbf{O}\in\mathbb{R}^{N\times d}$ to the image features through cross-attention. The queries $\mathbf{O}$ are used as prompts in DETR for closed-set object detection. To accommodate for open-set FSOD, we propose to provide 
novel classes' templates  
as additional \textit{visual prompts} in order to condition and control the detector's output. To train the system, we also propose to ``stamp'' these prompts with \textit{pseudo-class embeddings}, akin to soft-prompting, which are then predicted by the decoder along with bounding boxes. This can be viewed as an analogous component to the traditional positional embeddings.
The proposed \mname~ is depicted in Fig.~\ref{fig:overall_idea}. Compared to~\cite{carion2020end}, we highlight \textcolor{carnelian}{key differences} in our mathematical formulation in \textcolor{carnelian}{red}.

\subsection{\mname}~\label{ssec:\mname-arch} 
The following subsections detail \mname's architecture and main components. 

\noindent\textbf{Template encoding:} Let $\mathbf{T}_{i,j}\in \mathbb{R}^{H_p \times W_p \times 3}, i\in \{1,\dots,m\}, j\in\{1,\dots,k\}$
be the template images of the available classes (sampled from $C_{base}$ during training) where $m$ is the number of classes at the current training iteration ($m$ can vary), and $k$ is the number of examples per class (\ie k-shot detection; $k$ can also vary). A CNN backbone (\eg ResNet-50) generates template features $\mathbf{X} = \texttt{CNN}(\mathbf{T}),\;\mathbf{X}\in\mathbb{R}^{mk \times d }$ using either average or attention pooling (see Sec.~\ref{ssec:template_encoder}). 
% \\

\noindent\textbf{Pseudo-class embeddings:} We propose to dynamically and randomly associate, at each training iteration, the $k$ template prompts in $\mathbf{X}$ belonging to the i-th class (for that iteration) with a pseudo-class represented by a pseudo-class embedding $\mathbf{c^s_i}\in\mathbb{R}^d$, which are added to the templates as follows:
\vspace{-0.3cm}
\begin{equation}
    \color{carnelian}\mathbf{X}^s =  \mathbf{X} + \mathbf{C}^s, 
    \label{eq:embeddings}
    \vspace{-0.1cm}
\end{equation}
where $\mathbf{C}^s\in \mathbb{R}^{mk \times d}$ contains the pseudo-class embeddings for all templates at the current iteration. The pseudo-class embeddings  are initialised from a normal distribution and learned during training. They are not determined by the ground-truth categories and are class-agnostic. During each inference step, we arbitrarily associate to a template prompt (belonging to some class) the $i\textrm{-th}$ embedding as described by Eq.~\ref{eq:embeddings}. The goal is to retrieve the pseudo-class $i$. Note that the actual class information is not used. As the assigned embedding changes at every iteration, there is no correlation between the actual classes and the learned embeddings. See also Fig.~\ref{fig:overall_idea} that exemplifies this process.
In the proposed \mname, each decoded object query $\mathbf{o}_i$ in $\mathbf{O}$ will attempt to predict a pseudo-class using a classifier.  
Pseudo-class embeddings
add a signature to each visual prompt allowing the network to track the template within and dissociate it from the rest of the templates belonging to a different class. As transformers are permutation invariant, this vectors are required in order to track the visual prompt within the model.

\noindent\textbf{Templates as visual prompts:} We propose to provide the templates $\mathbf{X}^s$ as \textit{visual prompts} to the system by prepending them to the sequence of object queries fed to the decoder:
\begin{equation}
    \color{carnelian}\mathbf{O}^{\prime} =  \large[\mathbf{X}^s \;\; \mathbf{O}\large],\;\; \mathbf{O}^{\prime}\in\mathbb{R}^{{(mk+N)}\times d}.  
\end{equation}
As shown below, the templates will induce pseudo-class related information into the object queries via attention. This can be interpreted as a new form of training-aware soft-prompting~\cite{liu2021pre}. 

\noindent\textbf{\mname~ encoder:} Given a target image $\mathbf{I}\in\mathbb{R}^{H' \times W' \times 3}$, the same CNN backbone used for template feature extraction 
first
generates image features $\mathbf{Z} = \texttt{CNN}(\mathbf{I}),\;\mathbf{Z}\in\mathbb{R}^{S \times d },\; S=H \times W$, which are enriched with positional information through positional encodings $\mathbf{Z} \leftarrow \mathbf{Z} + \mathbf{P}_{s},\; \mathbf{P}_{s}\in\mathbb{R}^{S\times d}$. The features $\mathbf{Z}$ are then processed by \mname's encoder layers in order to be enriched with global contextual information. 
The $l-$th encoding layer processes the output features of the previous layer $\mathbf{Z}^{l-1}$ using a series of Multi-Head Self-Attention (MHSA), Layer Normalization (LN), and MLP layers (typical in~\cite{vaswani2017attention} and~\cite{carion2020end}), as well as a newly proposed Multi-Head Cross-Attention (MHCA) layer as follows~\footnote{
We follow DETR's notation
where $\mathbf{O'}$ is added to $LN(\mathbf{V}^{l-1})$ and then projected to form the query Q and key K for self-attention. 
Here, the first layer $\mathbf{V}^{l-1}$ is initialised as $[X_s, zeros]$ while in DETR with $zeros$.
}:
\vspace{-0.2cm}
\begin{eqnarray}
\label{eq:cross-encoder-1}
\mathbf{Z}^{\prime} & = & \textrm{MHSA}(\textrm{LN}(\mathbf{Z}^{l-1})) + \mathbf{Z}^{l-1},\\
\label{eq:cross-encoder-2}
\color{carnelian} \mathbf{Z}^{\prime\prime} & \color{carnelian}= & \color{carnelian}\textrm{MHCA}(\textrm{LN}(\mathbf{Z}^{\prime}), \mathbf{X}^s) + \mathbf{Z}^{\prime},\\
\label{eq:cross-encoder-3}
\mathbf{Z}^{l} & = & \textrm{MLP}(\textrm{LN}(\mathbf{Z}^{\prime\prime})) + \mathbf{Z}^{\prime\prime}.
\end{eqnarray}
The purpose of the MHCA layer above is to filter and highlight early on, before decoding, the image tokens of interest. We have found that such a layer noticeably increases few-shot accuracy (see also Section~\ref{sec:ablation}).
\mname's encoder is implemented by stacking $L=6$ blocks, each following Eq.~(\ref{eq:cross-encoder-1})-(\ref{eq:cross-encoder-3}). As image tokens are permutation invariant, we followed~\cite{carion2020end} and used a fixed positional encoding. For the templates, pseudo-class embeddings serve as positional encodings.
% \\

\noindent\textbf{\mname~ decoder:} \mname's decoder accepts as input the concatenated templates and learnable object queries  $\mathbf{O}^{\prime}$ which are transformed
by the decoder's layers through self-attention and cross-attention layers in order to be eventually used for pseudo-class prediction and bounding box regression. The $l-$th decoding layer processes the output features of the previous layer $\mathbf{V}^{l-1}$ as follows:
\vspace{-0.3cm}
\begin{eqnarray}
\label{eq:decoder-1}
\mathbf{V}^{\prime} & = & \textrm{MHSA}(\textrm{LN}(\mathbf{V}^{l-1})+\color{carnelian}\mathbf{O}^{\prime}\color{black}) + \mathbf{V}^{l-1},\\
\label{eq:decoder-2}
\mathbf{V}^{\prime\prime} & = & \textrm{MHCA}(\textrm{LN}(\mathbf{V}^{\prime})+\color{carnelian}\mathbf{O}^{\prime}\color{black}, \mathbf{Z}^{l}) + \mathbf{V}^{\prime},\\
\label{eq:decoder-3}
\mathbf{V}^{l} & = & \textrm{MLP}(\textrm{LN}(\mathbf{V}^{\prime\prime})) + \mathbf{V}^{\prime\prime},
\label{eq:decoder}
\end{eqnarray}
% where $\mathbf{V}^{0}=\texttt{zeros}(mk+N, d)$. 
where $\mathbf{V}^{0} =\large[\mathbf{X}^s \;\; \texttt{zeros}(N, d)\large]$.
Notably, different MLPs are used to process the decoder's features $\mathbf{V}=[\mathbf{V}_{\mathbf{X}^s}\;\mathbf{V}_{\mathbf{O}}]$ corresponding to the templates $\mathbf{V}_{\mathbf{X}^s}$ and the object queries $\mathbf{V}_{\mathbf{O}}$:
\begin{equation} 
 \color{carnelian}\large\texttt{MLP}(\mathbf{V}) = \large[\texttt{MLP}(\mathbf{V}_{\mathbf{X}^s}) \;\; \texttt{MLP}(\mathbf{V}_{\mathbf{O}})\large]. 
 \label{eq:decoder-mlp}
\vspace{-0.2cm}
\end{equation}
\mname's decoder consists of $L=6$ layers implemented using Eqs.~(\ref{eq:decoder-1})-(\ref{eq:decoder-mlp}).
% \\

\noindent\textbf{\mname~ training and loss functions:} For each base class that exists in the target image, we create a template for that class by randomly sampling and cropping an object from that category using a different image (containing an object of the same class) from the train set. After applying image augmentation, the cropped object/template is passed through the CNN backbone of \mname. For each target image and template $i$ (depicted in that image), the ground truth is $y_i= (c^s_i,b_i)$,  where $c^s_i$ is the target  pseudo-class label (up to $m$ classes in total) and $b_i\in[0,1]^4$ are the normalised bounding box coordinates. To calculate the loss for training \mname, only the $N$ transformed object queries $\mathbf{V}^L_{\mathbf{O}} \in \mathcal{R}^{N\times d}$ from the output of the last decoding layer are used for pseudo-class classification and bounding box regression (\ie $\mathbf{V}^L_{\mathbf{X}^s}$ is not used). To this end, pseudo-class and bounding box prediction heads are used to produce a set of $N$ predictions $ \{\hat{y}_{i}\}_{i=1}^{N}$ consisting of the pseudo-class probabilities $\hat{p}_i(c^s)$ and the bounding box coordinates $\hat{b}_{i}$. The heads are implemented using a 3-layer MLP and ReLU activations. Similarly to~\cite{carion2020end}, we used an additional special pseudo-class $\varnothing$ to denote tokens without valid object predictions. 
Note that as the training is done in a class-agnostic way via mapping of the base class templates to pseudo-classes (the actual class information is discarded) the model is capable to generalise to the unseen novel categories. %A2.1 }

Following~\cite{carion2020end}, 
bipartite matching is used to find an optimal permutation $\{\hat{y}_{\sigma{i}}\}_{i=1}^{N}$. Finally, the loss is:
\vspace{-0.2cm}
\begin{dmath}
L = \sum_{i=1}^N \color{carnelian}\lambda_1 L_{\rm{CE}}(c^s_i, \hat{p}_{\sigma(i)}(c^s)) \color{black} + \lambda_2 ||b_i - \hat{b}_{\sigma(i)}||_1  + \lambda_3 \texttt{IoU}(b_i, \hat{b}_{\sigma(i)}),
\end{dmath}\label{eq:loss}
where $\texttt{IoU}$ is the GIoU loss of~\cite{rezatofighi2019generalized} and $\lambda_i$ are the loss weights.

\noindent\textbf{Pre-training:} Transformers are generally \textit{more data hungry} than CNNs due to their lack of inductive bias~\cite{dosovitskiy2020image}. Therefore, 
% unsurprisingly, 
building representations that generalise well to unseen data, and prevent overfitting within the DETR framework, requires larger amounts of data. To this end, we used images from ImageNet-100~\cite{tian2019contrastive} and to some extent MSCOCO, for unsupervised pre-training where the classes and the bounding boxes are generated on-the-fly using an object proposal system, \textit{without using any labels}. \adrian{Note, that unlike all prior works, we make use of neg. templates as prompts, training the network using our proposed loss.} See also the appendix.

%% file: sections/experiments.tex
\paragraph{\bf Datasets:}
Experiments presented 
here
were all conducted using PASCAL VOC~\cite{everingham2010pascal,voc2012} and MSCOCO~\cite{coco2014} datasets. Moreover, ImageNet100~\cite{tian2019contrastive}, consisting of $\sim$125K images and 100 categories, is used (without labels) to pre-train our object detector. PASCAL VOC and MSCOCO are used to train and evaluate few-shot experiments. Following previous works~\cite{kang2019few,wang2020frustratingly,han2021query}, we evaluate 
the proposed method 
on PASCAL VOC 2007+2012 and MSCOCO 2014, using the same data splits provided by \cite{kang2019few,wang2020frustratingly}.
Specifically, PASCAL VOC is 
randomly 
divided into three different splits, each consisting of 15 base and 5 novel classes; 
training is done on the PASCAL VOC 2007+2012 train/val sets, and evaluation on the PASCAL VOC 2007 test set. 
Similarly, MSCOCO is split into base and novel categories, where the 20 overlapping categories with PASCAL VOC are considered novel, while the remainder are the base categories; 
following recent convention~\cite{kang2019few,wang2020frustratingly,han2021query}, 5k samples from the validation set are held out for testing, while the remaining samples from both train and validation sets are used for training.

\noindent{\bf Evaluation setting:} There are currently two widely-used FSOD evaluation protocols. The first focuses exclusively on 
% the performance on 
novel classes while disregarding base class performance, thus not monitoring 
catastrophic forgetting. The second, more comprehensive protocol, 
called generalised few-shot object detection (G-FSOD), 
considers
both base and novel classes. The choice of protocol and, hence, results interpretation, bears special importance for re-training based methods, as 
base class generalizability might be compromised. 
Without re-training methods, as \mname, adhere to the second protocol (G-FSOD) by default, as
base class catastrophic forgetting is not applicable.
\ric{As in~\cite{wang2020frustratingly}, we report results from several runs using different templates, and hence, report competing results using a likewise setting when available.}

\noindent{\bf Baselines:} Existing FSOD methods can be broadly categorised 
into:

re-training based, and 
% 2) 
without re-training. The latter can handle few-shot detection on the fly at 
deployment, while 
re-training based FSOD methods generally tend to perform better. Re-training based methods can be further subdivided into ``meta-learning'' and ``fine-tuning'' approaches. 
\noindent\textit{``Re-training based: meta-learning''} approaches include:  
Xiao \etal~\cite{xiao2020few}, 
DCNET~\cite{hu2021dense}, 
TIP~\cite{li2021transformation} and 
QA-FewDet~\cite{han2021query}.
\noindent\textit{``Re-training based: fine-tuning''} methods include:
Fan \etal~\cite{fan2020few}, CME~\cite{li2021beyond}, SRR-FSD~\cite{zhu2021semantic}, Zhang \etal~\cite{zhang2021hallucination}, DeFRCN~\cite{qiao2021defrcn},
\ric{FCT~\cite{han2022few}, KFSOD~\cite{zhang2022kfsod}, TENET~\cite{zhang2022tenet}, FSODMC~\cite{QiFan2022},}
\adrian{DETReg~\cite{bar2022detreg}, Meta-DETR~\cite{zhang2022meta} and tsf~\cite{lai2022tsf}}. 
\noindent\textit{``Without re-training''} methods include: UP-DETR~\cite{dai2021up}, Fan \etal~\cite{fan2020few}, QA-FewDet~\cite{han2021query},
\ric{Meta Faster R-CNN~\cite{han2022meta}} and \adrian{AirDet~\cite{li2022airdet}}. Note that these \ric{4} last methods can also be re-trained, offering improved accuracy.

\subsection{Overview of Results}
From our results below on both datasets, we can take away two messages:
\vspace{-0.15cm}
\begin{itemize}[leftmargin=*]
\itemsep0em 
\item 
\textbf{Conclusion 1:} \mname~outperforms \textbf{all} without re-training based approaches by a large margin, 
\ie those directly comparable. 
\item 
\textbf{Conclusion 2:} \mname~ 
outperforms the majority of re-training based approaches (some by a large margin). 
Importantly, on average across all novel sets from PASCAL VOC, it outperforms \textbf{all} 
for $k=1$, while at the same time being more robust across splits, \ie \mname~ has lower variance across novel sets. Similarly, in MSCOCO with $k=1,2$ it outperforms nearly all re-trained methods.
\item 
\textbf{Conclusion 2:} \mname~ Shows strong and conistent performance across a variable number of shots.
\end{itemize}
\vspace{-0.3cm}

\begin{table*}[t]
\centering
\footnotesize
\setlength{\tabcolsep}{0.4em}
\caption{FSOD performance (AP50) on the PASCAL VOC dataset. Results with $^{\dag}$ are from~\cite{han2021query} while those with $^\ddag$ were produced by us. 
Our approach outperforms all without  re-training methods. Moreover, it provides competitive results compared with other re-training based methods for $k=3,5,10$, and even outperforms \ric{most} for $k=1,2$, \ie extreme few-shot settings.
}
%\adjustbox{width=\linewidth}{
\begin{tabular}{l|c|c|ccccc|ccccc|ccccc}
\toprule
\multirow{2}{*}{Method / Shot} & \multirow{2}{*}{Venue} & \multirow{2}{*}{Backbone} & \multicolumn{5}{c|}{Novel Set 1} & \multicolumn{5}{c|}{Novel Set 2} & \multicolumn{5}{c}{Novel Set 3} \\ 
&  & & 1     & 2     & 3    & 5    & 10   & 1     & 2     & 3    & 5    & 10   & 1     & 2     & 3    & 5    & 10   \\  \midrule
% \multicolumn{18}{c}{\textbf{Fine-tuning the model on novel classes, and testing on novel classes}} \\ \midrule
\multicolumn{18}{c}{\textbf{Re-training based methods (meta-learning or fine-tuning)}} \\ \midrule
 FSRW$^{*}$~\cite{kang2019few}  & ICCV'19 & YOLOv2 & 14.8  & 15.5  & 26.7 & 33.9 & 47.2 & 15.7  & 15.3  & 22.7 & 30.1 & 40.5 & 21.3  & 25.6  & 28.4 & 42.8 & 45.9 \\ 
 MetaDet$^{*}$ ~\cite{wang2019meta}& ICCV'19 & VGG16 & 18.9 & 20.6 & 30.2 & 36.8 & 49.6 & 21.8 & 23.1 & 27.8 & 31.7 & 43.0 & 20.6 & 23.9 & 29.4 & 43.9 & 44.1 \\ 
 Meta R-CNN$^{*}$~\cite{yan2019meta} & ICCV'19 & RN-101 & 19.9 & 25.5 & 35.0 & 45.7 & 51.5 & 10.4 & 19.4 & 29.6 & 34.8 & 45.4 & 14.3 & 18.2 & 27.5 & 41.2 & 48.1 \\ 
 TFA w/fc \cite{wang2020frustratingly} & ICML'20 & RN-101 & {36.8} & {29.1} & {43.6} & {55.7} & {57.0} & {18.2} & {29.0} & {33.4} & {35.5} & {39.0} & {27.7} & {33.6} & {42.5} & {48.7} & {50.2}\\
TFA w/cos~\cite{wang2020frustratingly}  & ICML'20 & RN-101 & 25.3 & 36.4 & 42.1 & 47.9 & 52.8 & 18.3 & 27.5 & 30.9 & 34.1 & 39.5 & 17.9 & 27.2 & 34.3 & 40.8 & 45.6  \\ 
 TFA w/cos$^{*}$~\cite{wang2020frustratingly} & ICML'20 & RN-101 & 39.8 & 36.1 & 44.7 & 55.7 & 56.0 & 23.5 & 26.9 & 34.1 & 35.1 & 39.1 & 30.8 & 34.8 & 42.8 & 49.5 & 49.8 \\ 
 Xiao \etal~\cite{xiao2020few} & ECCV'20 & RN-101 & 24.2 & 35.3 &  42.2 &  49.1 &  57.4 & 21.6 & 24.6 &  31.9 &  37.0 &  45.7 & 21.2 &  30.0 &  37.2 &  43.8 &  49.6 \\
 MPSR$^{*}$~\cite{wu2020multi} & ECCV'20 & RN-101 & 41.7 & 42.5 & 51.4 & 55.2 & 61.8 & 24.4 & 29.3 & 39.2 & 39.9 & 47.8 & 35.6 & 41.8 & 42.3 & 48.0 & 49.7 \\ 
Fan \textit{et al.}$^{\dag}$~\cite{fan2020few} & CVPR'20 & RN-101 & 37.8 & 43.6 & 51.6 & 56.5 & 58.6    & 22.5 & 30.6 & 40.7 & 43.1 & 47.6    & 31.0 & 37.9 & 43.7 & 51.3 & 49.8 \\
DCNET~\cite{hu2021dense} & CVPR'21 & RN-101 & 33.9 & 37.4 & 43.7 & 51.1 & 59.6 & 23.2 & 24.8 & 30.6 & 36.7 & 46.6 & 32.3 & 34.9 & 39.7 & 42.6 & 50.7 \\
TIP~\cite{li2021transformation} & CVPR'21 & RN-101 & 27.7 & 36.5 & 43.3 & 50.2 & 59.6 & 22.7 & 30.1 & 33.8 & 40.9 & 46.9 & 21.7 & 30.6 & 38.1 & 44.5 & 50.9 \\
CME~\cite{li2021beyond} & CVPR'21 & RN-101 & 41.5 & 47.5 & 50.4 & 58.2 & 60.9 & 27.2 & 30.2 & 41.4 & 42.5 & 46.8 & 34.3 & 39.6 & 45.1 & 48.3 & 51.5 \\
SRR-FSD~\cite{zhu2021semantic} & CVPR'21 & RN-101 & 47.8 &  50.5 & 51.3 & 55.2 & 56.8 & 32.5 & 35.3 & 39.1 & 40.8 & 43.8 & 40.1 & 41.5 & 44.3 & 46.9 & 46.4 \\
Zhang \textit{et al.}$^{*}$~\cite{zhang2021hallucination} & CVPR'21 & RN-101 & 47.0 & 44.9 & 46.5 & 54.7  & 54.7 & 26.3 & 31.8 & 37.4 & 37.4 & 41.2 & 40.4 & 42.1 & 43.3 & 51.4 & 49.6 \\
QA-FewDet~\cite{han2021query} & ICCV'21 & RN-101 & 42.4 & 51.9 & 55.7 & 62.6 & 63.4 & 25.9 & 37.8 & 46.6 & 48.9 & 51.1 & 35.2 & 42.9 & 47.8 & 54.8 & 53.5 \\
DeFRCN~\cite{qiao2021defrcn} & ICCV'21 & RN-101 & 40.2 & 53.6 & 58.2 & 63.6 & 66.5 & 29.5 & 39.7 & 43.4 & 48.1 & 52.8 & 35.0 & 38.3 & 52.9 & 57.7 & 60.8 \\
\rowcolor{tabhighlight}
DeFRCN+TSF~\cite{lai2022tsf} & ECCV'22 & RN-101 & 43.6 & 57.4 & 61.2 & 65.1 & 65.9 & 31.0 & 40.3 & 45.3 & 49.6 & 52.5 & 39.3 & 51.4 & 54.8 & 59.8 & 62.1\\
DeFRCN$^*$~\cite{qiao2021defrcn} & ICCV'21 & RN-101 & 53.6 & 57.5 & 61.5 & 64.1 & 60.8 & 30.1 & 38.1 & 47.0 & 53.3 & 47.9 & 48.4 & 50.9 & 52.3 & 54.9 & 57.4 \\
\rowcolor{tabhighlightric}
FCT~\cite{han2022few} & CVPR'22 & PVTv2-B2-Li & 38.5 & 49.6 & 53.5 & 59.8 & 64.3 & 25.9 & 34.2 & 40.1 & 44.9 & 47.4 & 34.7 & 43.9 & 49.3 & 53.1 & 56.3 \\
\rowcolor{tabhighlightric}
KFSOD~\cite{zhang2022kfsod} & CVPR'22 & RN-50 & 44.6 & - & 54.4 & 60.9 & 65.8 & 37.8 & - & 43.1 & 48.1 & 50.4 & 34.8 & - & 44.1 & 52.7 & 53.9 \\
\rowcolor{tabhighlightric}
TENET~\cite{zhang2022tenet} & ECCV'22 & RN-50 & 46.7 & - & 55.4 & 62.3 & 66.9 & 40.3 & - & 44.7 & 49.3 & 52.1 & 35.5 & - & 46.0 & 54.4 & 54.6 \\
\rowcolor{tabhighlightric}
% FSODMC~\cite{QiFan2022} & ECCV'22 & RN-50 & 40.1 & 44.2 & 51.2 & 62.0 & 63.0 & 33.3 & 33.1 & 42.3 & 46.3 & 52.3 & 36.1 & 43.1 & 43.5 & 52.0 & 56.0 \\
\rowcolor{tabhighlightric}
Meta Faster R-CNN~\cite{han2022meta} & AAAI'22 & RN-101 & 43.0 & 54.5 & 60.6 & 66.1 & 65.4 & 27.7 & 35.5 & 46.1 & 47.8 & 51.4 & 40.6 & 46.4 & 53.4 & 59.9 & 58.6 \\
\rowcolor{tabhighlight}
% FCT~\cite{han2022few} & CVPR'22 & PVTv2-B2-Li & 38.5 & 49.6 & 53.5 & 59.8 & 64.3 & 25.9 & 34.2 & 40.1 & 44.9 & 47.4 & 34.7 & 43.9 & 49.3 & 53.1 & 56.3 \\
\rowcolor{tabhighlight}
Meta-DETR~\cite{zhang2022meta} & TPAMI'22 & DETR-R101 &  35.1 & 49.0 & 53.2 & 57.4 & 62.0 & 27.9 & 32.3 & 38.4 & 43.2 & 51.8 & 34.9 & 41.8 & 47.1 & 54.1 & 58.2\\
\midrule
% \multicolumn{18}{c}{\textbf{Meta-training the model on base classes, and meta-testing on novel classes}} \\ \midrule
\multicolumn{18}{c}{\textbf{Without re-training methods}} \\ \midrule
Fan \textit{et al.}$^{\dag}$~\cite{fan2020few} & CVPR'20 & RN-101 & 32.4 & 22.1 & 23.1 & 31.7 & 35.7    & 14.8 & 18.1 & 24.4 & 18.6 & 19.5 &    25.8 & 20.9 & 23.9 & 27.8 & 29.0 \\
UP-DETR$^\ddag$~\cite{dai2021up} & ICCV'21 & DETR-R50 & 38.2 & 40.4 & 44.5 & 45.8 & 46.0 & 20.0 & 23.6 & 25.8  & 28.0   & 33.9 & 34.1 & 35.3 & 37.0 & 40.1 & 40.3 \\
QA-FewDet~\cite{han2021query} & ICCV'21 & RN-101 & 41.0 & 33.2 & 35.3 & 47.5 & 52.0   & 23.5 & 29.4 & 37.9 & 35.9 & 37.1   & 33.2 & 29.4 & 37.6 & 39.8 & 41.5 \\
\rowcolor{tabhighlightric}
Meta Faster R-CNN~\cite{han2022meta} & AAAI'22 & RN-101 & 40.2 & 30.5 & 33.3 & 42.3 & 46.9 & 26.8 & 32.0 & 39.0 & 37.7 & 37.4 & 34.0 & 32.5 & 34.4 & 42.7 & 44.3 \\
    \mname~(Ours) & this work & DETR-R50 & \textbf{45.0} & \textbf{48.5} & \textbf{51.5} & \textbf{52.7} & \textbf{56.1} & \textbf{37.3} & \textbf{41.3} & \textbf{43.4} & \textbf{46.6} & \textbf{49.0} & \textbf{43.8} & \textbf{47.1} & \textbf{50.6} & \textbf{52.1} & \textbf{56.9} \\
\bottomrule
\end{tabular}%}
\label{tab:main_voc}
\vspace{-0.35cm}
\end{table*}

\subsection{Results on PASCAL VOC}
\label{ssec:voc_results}
Table~\ref{tab:main_voc} summarises our results and compares them with 
the current state-of-the-art on PASCAL VOC.
Experiments for k-shot detection were conducted for three data splits, where k was set to ${1,2,3,4,5,10}$ and AP50 values are reported.
Note that Table~\ref{tab:main_voc} is split into two sections: 
Methods at the top require an additional few-shot re-training stage, while those at the bottom, 
including our method, do not require any re-training.
Here,
it can be appreciated that our approach outperforms all without re-training methods by a large margin, improving the current state-of-the-art~\cite{han2021query,dai2021up} in any shot and all split experiments by up to 17.8 AP50 points, and, in most cases, by at least $\sim$10 AP50 points. Moreover, and contrary to~\cite{han2021query,han2022meta}, our method can process multiple novel classes in a single forward pass. Finally, we re-implemented UP-DETR~\cite{dai2021up} 
for few-shot detection on PASCAL VOC (since there is no publicly available implementation for few-shot detection or results). Our method largely outperforms it, perhaps unsurprisingly, as the latter was not developed for few-shot detection, but for unsupervised pre-training.
Importantly, the proposed method provides competitive results or even outperforms re-training based methods (meta-learned or fine-tuned), \adrian{especially for extreme low-shot, $k=1,2$} (\eg ~\cite{li2021transformation,li2021beyond,zhu2021semantic,qiao2021defrcn,zhang2022meta,lai2022tsf}).
Qualitative visualizations in appendix.

\subsection{Results on MSCOCO}
\label{ssec:coco_results}

Table~\ref{tab:main_coco} shows evaluation results for \mname~ and all competing state-of-the-art methods on MSCOCO. 
Similarly to above,
Table~\ref{tab:main_coco} is split into methods requiring re-training at the top and those that do not require re-training at the bottom. There, it can be appreciated that \mname~ outperforms all comparable state-of-the-art methods~\cite{han2021query,fan2020few,li2022airdet} by up to $3.1$ AP50 points (1-shot) and, in most cases, by at least $\sim1.1$ AP50 points. In our experiments UP-DETR failed to converge on MSCOCO, hence, results are not included.
We speculate that this might be due to UP-DETR's partitioning the input queries by the number query patches, therefore, limiting the number of tokens query patches interact with. This appears to be too restrictive for MSCOCO.
Moreover, and in line with results observed on PASCAL VOC, \mname~ achieves competitive results to those of re-trained based methods on MSCOCO, a far more challenging dataset. \mname~ outperforms nearly all re-training based methods,  
for $k=1,2$  while performing comparably for $k=3,5,10$. This in itself is a promising results given our methods doesn't require retraining.

\begin{table*}[ht]
\centering
\footnotesize
\setlength{\tabcolsep}{0.4em}
\caption{FSOD performance on the MSCOCO dataset. 
Results with $^{\dag}$ are from~\cite{han2021query}.
Our method consistently outperforms the
state-of-the-art methods in most of the shots and metrics.\vspace{1mm}}
%\adjustbox{width=\linewidth}{
\begin{tabular}{l|ccc|ccc|ccc|ccc|ccc}
\toprule
&\multicolumn{3}{c|}{1-shot} & \multicolumn{3}{c|}{2-shot} & \multicolumn{3}{c|}{3-shot} &\multicolumn{3}{c|}{5-shot} & \multicolumn{3}{c}{10-shot}  \\
Method & AP & AP50 & AP75 & AP & AP50 & AP75 & AP & AP50 & AP75 & AP & AP50 & AP75 & AP & AP50 & AP75 \\%& AP & AP50 & AP75 \\ 
\midrule
% \multicolumn{18}{c}{\textbf{Fine-tuning the model on novel classes, and testing on novel classes}} \\ \midrule
\multicolumn{15}{c}{\textbf{Re-trained methods (meta-learning or fine-tuning)}} \\ \midrule
FSRW$^{*}$ \small{~\cite{kang2019few}} & {--} & {--} & {--} & {--} & {--} & {--} & {--} & {--} & {--} &\;{--} & {--} & {--} & 5.6 & 12.3 & 4.6 \\% & 9.1 & 19.0 & 7.6 \\ 
MetaDet$^{*}$\small{~\cite{wang2019meta}} & {--} & {--} & {--} & {--} & {--} & {--} & {--} & {--} & {--} &\;{--} & {--} & {--} & 7.1 & 14.6 & 6.1 \\% & 11.3 & 21.7 & 8.1 \\
Meta R-CNN$^{*}$~\cite{yan2019meta} & {--} & {--} & {--} & {--} & {--} & {--} & {--} & {--} & {--} &\;{--} & {--} & {--} & {8.7} & 19.1 & {6.6}  \\% & {12.4} & 25.3 & {10.8} \\
TFA w/fc \cite{wang2020frustratingly}$^{\dag}$ & 2.9 & 5.7 & 2.8   & 4.3 & 8.5 & 4.1    & 6.7 & 12.6 & 6.6   & 8.4 & 16.0 & 8.4    & 10.0 & 19.2 & 9.2  \\%  & 13.4 & 24.7 & 13.2 \\
TFA w/cos~\cite{wang2020frustratingly}           & 1.9 & 3.8 & 1.7   & 3.9 & 7.8 & 3.6    & 5.1 &  9.9 & 4.8   & 7.0 & 13.3 & 6.5     &  9.1 & 17.1 & 8.8  \\%  & 13.7& 24.9 & 13.4 \\
TFA w/cos$^{*\dag}$~\cite{wang2020frustratingly} & 3.4 & 5.8 & 3.8   & 4.6 & 8.3 & 4.8    & 6.6 & 12.1 & 6.5   & 8.3 & 15.3 & 8.0     & 10.0 & 19.1 & 9.3  \\%  & 13.7& 24.9 & 13.4 \\
Xiao \textit{et al.}$^{\dag}$~\cite{xiao2020few} & 3.2 & 8.9 & 1.4  & 4.9 & 13.3 & 2.3  & 6.7 & 18.6 & 2.9  & 8.1 & 20.1 & 4.4 & 10.7 & 25.6 & 6.5\\% & 15.9 & 31.7 & 15.1 \\
Fan \textit{et al.}$^{\dag}$~\cite{fan2020few} & 4.2 & 9.1 & 3.0   & 5.6 & 14.0 & 3.9   & 6.6 & 15.9 & 4.9   & 8.0 & 18.5 & 6.3   & 9.6 & 20.7 & 7.7   \\% & 13.5 & 28.5 & 11.7 \\
% DCNET~\cite{hu2021dense} & - & - & -  & - & - & - & - & - & - & - & - & - & 12.8 & 23.4 & 11.2 \\%& 18.6 &  32.6 & 17.5  \\ 
TIP~\cite{li2021transformation} & - & - & -  & - & - & - & - & - & - & - & - & - & 16.3 & 33.2 & 14.1 \\%& 18.3 & 35.9 & 16.9 \\
CME~\cite{hu2021dense} & - & - & -  & - & - & - & - & - & - & - & - & - & 15.1 & 24.6 & 16.4 \\%& 18.6 &  32.6 & 17.5  \\ 
SRR-FSD~\cite{zhu2021semantic} & - & - & -  & - & - & - & - & - & - & - & - & - & 11.3 & 23.0 & 9.8 \\%& 14.7 & 29.2 & 13.5 \\
Zhang \etal~\cite{zhang2021hallucination} & 4.4 & 7.5 & 4.9 & 5.6 & 9.9 & 5.9 & 7.2 & 13.3 & 7.4  & - & - & - & - & - & - \\%& - & - & - \\
QA-FewDet~\cite{han2021query} & 4.9 & 10.3 & 4.4  & 7.6 & 16.1 & 6.2    & 8.4 & 18.0 & 7.3  & 9.7 & 20.3 & 8.6   & 11.6 & 23.9 & 9.8 \\% & 16.5 & 31.9 & 15.5 \\
DeFRCN~\cite{qiao2021defrcn} & 4.8 & - & - & 8.5 & - & - & 10.7 & - & - & 13.6 & - & - & 16.8 & - & - \\  % & \\%21.2 & - & -  \\
\rowcolor{tabhighlight}
DeFRCN+TSF~\cite{lai2022tsf} & 5.0 & - & - & 8.7 & - & - & 10.9 & - & - & 13.6 & - & - & 16.6 & - & - \\
% DeFRCN$^{*}$~\cite{qiao2021defrcn}  & 9.3 & - & - & 12.9 & - & - & 14.8 & - & - & 16.1 & - & - & 18.5 & - & -   \\%& 22.6 & - & -  \\
\rowcolor{tabhighlight}
DetReg~\cite{bar2022detreg} & - & - & - & - & - & - & - & - & - & - & - & - & 25.0 & - & 27.6 \\
\rowcolor{tabhighlightric}
FCT~\cite{han2022few} & 5.1 & - & - & 7.2 & - & - & 9.8 & - & - & 12.0 & - & - & 15.3 & - & - \\
\rowcolor{tabhighlightric}
KFSOD~\cite{zhang2022kfsod} & - & - & - & - & - & - & - & - & - & - & - & - & - & 25.7 & 14.6 \\
\rowcolor{tabhighlightric}
FSODMC~\cite{QiFan2022} & - & - & - & - & - & - & - & - & - & 15.1 & 27.2 & 14.6 & - & - & - \\
\rowcolor{tabhighlightric} 
Meta Faster R-CNN~\cite{han2022meta} &  5.1 & 10.7 & 4.3 & 7.6 & 16.3 & 6.2 & - & - & - & - & - & - & 12.7 & 25.7 & 10.8 \\
\rowcolor{tabhighlight}
FCT~\cite{han2022few} & 5.1 & - & - & 7.2 & - & - & 9.8 & - & - & 12.0 & - & - & 15.3 & - & - \\
\rowcolor{tabhighlight}
Meta-DETR~\cite{zhang2022meta}& 7.5 & 12.5 & 7.7 & - & - & - & 13.5 & 21.7 & 14.0 & 15.4 & 25.0 & 15.8  & 19.0 & 30.5 & 19.7 \\
\rowcolor{tabhighlight}
AirDet~\cite{li2022airdet} & 6.10 & 11.40 & 6.04 & 8.73 & 16.24 & 8.35 & 9.95 & 19.39 & 9.09 & 10.81 & 20.75 & 10.27 & 13.0 & 23.9 & 12.4  \\
\midrule
% \multicolumn{19}{c}{\textbf{Meta-training the model on base classes, and meta-testing on novel classes}} \\ \midrule
\multicolumn{15}{c}{\textbf{Methods without re-training}} \\ \midrule
Fan \textit{et al.}$^{\dag}$~\cite{fan2020few} & 4.0 & 8.5 & 3.5  & 5.4 & 11.6 & 4.6    & 5.9 & 12.5 & 5.0   & 6.9 & 14.3 & 6.0    & 7.6 & 15.4 & 6.8  \\%& 8.9 & 17.8 & 8.0 \\
QA-FewDet~\cite{han2021query} &  5.1 & 10.5 & 4.5    & 7.8 & 16.4 & 6.6    & 8.6 & 17.7 & 7.5    & 9.5 & 19.3 & 8.5  & 10.2 & 20.4 & 9.0 \\
\rowcolor{tabhighlightric} 
Meta Faster R-CNN~\cite{han2022meta} & 5.0 & 10.2 & 4.6 & 7.0 & 13.5 & 6.4 & - & - & - & - & - & - & 9.7 & 18.5 & 9.0 \\
\rowcolor{tabhighlight}
AirDet~\cite{li2022airdet} & 5.97 & 10.52 & 5.98 & 6.58 & 12.02 & 6.33 & 7.00 & 12.95 & 6.71 & 7.76 & 14.28 & 7.31 & 8.7 & 15.3 & 8.8 \\
\mname~(Ours) & \textbf{7.0}  & \textbf{13.6}  & \textbf{7.5}    & \textbf{8.9}  & \textbf{17.5}  & \textbf{9.0}    & \textbf{10.0} & \textbf{18.8} & \textbf{10.0}  & \textbf{10.9}  & \textbf{20.7}  &  \textbf{10.8}  &  \textbf{11.3}  & \textbf{21.7}  & \textbf{11.1}  \\
\bottomrule
\end{tabular}%}
\label{tab:main_coco}
%\vspace*{-3mm}
\end{table*}
%\vspace{-0.25cm}
\begin{table}[ht]
\centering
\small
\setlength{\tabcolsep}{0.4em}
\caption{FSOD performance (AP50) on the PASCAL VOC dataset Novel Set 1 for various template construction configurations. $\dagger$ - result produced using 
% extra 
bounding-box jittering for the patch extraction.}
%\adjustbox{width=0.6\linewidth}{
\begin{tabular}{l|c|ccccc}
\toprule
\multirow{2}{*}{Resolution}  &  \multirow{2}{*}{Pool. type} &  \multicolumn{5}{c}{Novel Set 1} \\ 
& &    1     & 2     & 3    & 5    & 10     \\  \midrule
128 &  global.avg. & 42.9 & 46.0 & 49.4 & 50.5 & 54.0   \\
128 &  attn. & 45.0 & 48.5 & 51.5 & 52.7 & 56.1    \\
$128^\dagger$ &  attn. & 39.0 & 42.8 & 44.6 & 46.4 & 50.3   \\
96 &  attn. & 43.2 & 45.7 & 49.0 & 50.1 & 52.9   \\
192 &  attn. & 45.1 & 48.3 & 51.0 & 52.9 & 57.0   \\
\bottomrule
\end{tabular}%}
\label{tab:patch_comparison}
%\vspace*{-0.3cm}
\end{table}

%% file: sections/ablation.tex
Herein, we ablate different variations and components of our method, analysing the impact of different design choices. Unless otherwise specified, we report results on Novel Set 1 on PASCAL VOC. 
For more details 
see the appendix.

\noindent{\bf Design of the prompt template encoder:}
\label{ssec:template_encoder} An important component of our system is the extraction of discriminative prompts from the novel classes' templates.
To this end, we re-use \mname's input image CNN encoder.
However, to focus on the most important components we used attention-based pooling instead of simple global average pooling. In Table~\ref{tab:patch_comparison} we report the impact of: (a) resolution, (b) augmentation level, and (c) pooling type. As the results show, increasing the resolution from $128$ to $192$px yields no additional gains. This suggests that, at least for the datasets in question, fine grained details are not quintessential for the identification of the targeted novel class and higher level concepts suffice. While spatial augmentation generally helps (\ie for object recognition), we found that adding noise to the ground truth bounding box of the template at train time leads to lower accuracy. This makes the problem for the object detector too hard, and impedes convergence.
Finally, attentive pooling, instead of global average, can further boost performance. 

\noindent{\bf Pre-training:} Many FSOD systems use pre-trained backbones on ImageNet for classification. Deviating from this, 
% but within the same line of ideas, 
we pre-train our system in an unsupervised manner on ImageNet images and parts of MSCOCO without using the labels. We note that this is especially important for transformer based architectures which were shown to be more prone to over-fitting due to the lack of inductive bias~\cite{dosovitskiy2020image}. As the results from Table~\ref{tab:pretraining} show, unsupervised pre-training, can significantly boost the performance, preventing over-fitting toward the base classes and improving overall discriminative capacity. To reduce over-fitting the pre-training loss on ImageNet data is applied during supervised training every 8th iteration. 
Additional details and experiments in appendix.

% \begin{wraptable}[9]{r}{0.5\textwidth}
\begin{table}[ht]
\centering
\small
\setlength{\tabcolsep}{0.4em}
% \vspace{-0.7cm}
\caption{
% Few-shot object detection 
FSOD
performance (AP50) on the PASCAL VOC dataset on the Novel Set 1 for models with and without pre-training. }
% \adjustbox{width=0.9\linewidth}{
\begin{tabular}{c|ccccc}
\toprule
\multirow{2}{*}{Pre-training}  &  \multicolumn{5}{c}{Novel Set 1} \\ 
&   1     & 2     & 3    & 5    & 10     \\  \midrule
 & 19.0 &  21.1  & 23.3 & 24.0  & 24.6    \\
\checkmark  & 45.0 & 48.5 & 51.5 & 52.7 & 56.1   \\
\bottomrule
\end{tabular}%}
\vspace{-1.5em}
\label{tab:pretraining}
\end{table}

% \subsection{Auxiliary losses}
\noindent{\bf Auxiliary losses:} We explored the impact of using additional auxiliary losses applied to the object queries, an $L_2$ feature loss and a contrastive loss, where the positive pairs are formed by taking the input templates with all the object query tokens assigned to it by the Hungarian assignment algorithm. We did not observe any further gains from the additional losses, suggesting that the pseudo-classification loss alone suffices for guiding the network.

\noindent{\bf Impact on individual components:} Herein, we analyse the accuracy improvement obtained by two components of \mname~namely the MHCA layer in \mname's encoder (see Eq.~\ref{eq:cross-encoder-3}), and the type-specific MLPs (TS-MLP) in \mname's decoder (see Eq.~\ref{eq:decoder-mlp}). As Table~\ref{tab:individual_components} shows, while our system, without both components, provides satisfactory results, unsurprisingly, the addition of TS-MLP further boosts the accuracy. This is expected as the information carried by the object queries and template tokens is semantically different, so ideally they should be transformed using different functions. 
Finally, the MHCA within the encoder injects template-related information early on to filter or highlight certain 
image regions,
and also helps increase the accuracy.

\vspace{-0.1cm}
\begin{table}[ht]
\centering
\small
\setlength{\tabcolsep}{0.4em}
% \vspace{-0.1cm}
\caption{Impact of various components on the 
% few-shot object detection 
FSOD
performance (AP50) on the PASCAL VOC dataset (Novel Set 1). }
\adjustbox{width=0.99\linewidth}{
\begin{tabular}{c|ccccc}
\toprule
\multirow{2}{*}{Approach}  &  \multicolumn{5}{c}{Novel Set 1} \\ 
&   1     & 2     & 3    & 5    & 10     \\  \midrule
\mname~w/o TS-MLP & 42.2 & 46.9 & 48.3 & 49.2 & 51.6    \\
\mname~w/o MHCA of Eq.~\ref{eq:cross-encoder-2} & 38.1 & 40.6 & 41.7 & 42.2 & 45.6   \\
\mname~& 45.0 & 48.5 & 51.5 & 52.7 & 56.1   \\
\bottomrule
\end{tabular}}
\label{tab:individual_components}
% \vspace{-0.5cm}
\end{table}

%% file: sections/conclusions.tex
In this work we propose \mname, a novel transformer based few-shot architecture, that is simple, powerful and flexible. \mname~outperforms all prior training-free methods, thus achieving a new state-of-the-art.
In addition to the outstanding results presented in Sec.~\ref{sec:experiments}, the proposed method can simultaneously predict arbitrary number of classes, using variable-shots per class, in a single forward pass. These results, in combination with the methods formulation, clearly demonstrate not only its performance improvements but also its high flexibility. Therefore, the visual prompting framework proposed \mname~ can uniquely satisfy the outlined FSOD system \textit{desiderata} (a) and (b), while 
also
making big improvements toward satisfying (c). 